\documentclass{article}

\usepackage{PRIMEarxiv}

\usepackage[utf8]{inputenc} 
\usepackage[T1]{fontenc}    
\usepackage{hyperref}       
\usepackage{url}            
\usepackage{booktabs}       
\usepackage{amsfonts}       
\usepackage{nicefrac}       
\usepackage{microtype}      
\usepackage{lipsum}
\usepackage{fancyhdr}       
\usepackage{graphicx}       
\usepackage{caption}
\usepackage{tabulary} 
\usepackage[flushleft]{threeparttable}
\usepackage{multirow}
\usepackage{adjustbox}
\graphicspath{{media/}}     
\usepackage{subcaption}
\usepackage{amsmath,amsfonts,bm}
\usepackage[toc,page]{appendix}
\usepackage[colorinlistoftodos,prependcaption,textsize=tiny]{todonotes}
\usepackage{float}  
\usepackage{algorithm} 
\usepackage{algpseudocode} 
\usepackage{makecell}
\usepackage{hyperref}
\usepackage{ulem}
\usepackage{longtable}
\usepackage{array}

\newcommand{\lihanga}[1]{{\color{black}{#1}}} 
\newcommand{\fang}[1]{{\color{black}{#1}}} 

\title{Pre-Training on Large-Scale Generated Docking Conformations with HelixDock \\to Unlock the Potential \\of Protein-Ligand Structure Prediction Models}

\author{
 Lihang Liu$^1$\thanks{Equal contributions.}, Shanzhuo Zhang$^1$\footnotemark[1], Donglong He$^1$, Xianbin Ye$^1$, Jingbo Zhou$^1$, Xiaonan Zhang$^1$, \\
 \textbf{Yaoyao Jiang$^3$, Weiming Diao$^4$, Hang Yin$^{3,4}$, Hua Chai$^2$, Fan Wang$^1$, Jingzhou He$^1$, Liang Zheng$^2$,} \\
 \textbf{Yonghui Li$^2$\footnotemark[2], Xiaomin Fang$^1$\thanks{Corresponding authors. Email: fangxiaomin01@baidu.com and liyh@cdcszx.cn}} \\
 $^1$PaddleHelix team, Baidu Inc., \\
 $^2$National Supercomputing Center in Chengdu, \\
 $^3$School of Pharmaceutical Sciences, Key Laboratory of Bioorganic Phosphorous Chemistry and Chemical \\Biology (Ministry of Education), Tsinghua-Peking Center for Life Sciences, Tsinghua University, Beijing, China, \\
 $^4$Toll Biotech Co. Ltd (Beijing), Beijing, China
 }

\begin{document}
\maketitle

\begin{abstract}
Protein-ligand structure prediction is an essential task in drug discovery, predicting the binding interactions between small molecules (ligands) and target proteins (receptors). Although conventional physics-based docking tools are widely utilized, their accuracy is compromised by limited conformational sampling and imprecise scoring functions. Recent advances have incorporated deep learning techniques to improve the accuracy of protein-ligand structure prediction. Nevertheless, the experimental validation of docking conformations remains costly, it raises concerns regarding the generalizability of these deep learning-based methods due to the limited training data. In this work, we show that by pre-training on a large-scale docking conformation generated by traditional physics-based docking tools and then fine-tuning with a limited set of experimentally validated receptor-ligand complexes, we can obtain a protein-ligand structure prediction model with outstanding performance. Specifically, this process involved the generation of 100 million docking conformations for protein-ligand pairings, an endeavor consuming roughly 1 million CPU core days.  The proposed model, HelixDock, aims to acquire the physical knowledge encapsulated by the physics-based docking tools during the pre-training phase. HelixDock has been rigorously benchmarked against both physics-based and deep learning-based baselines, demonstrating its exceptional precision and robust transferability in predicting binding confirmation. In addition, our investigation reveals the scaling laws governing pre-trained protein-ligand structure prediction models, indicating a consistent enhancement in performance with increases in model parameters and the volume of pre-training data. Moreover, we applied HelixDock to several drug discovery-related tasks to validate its practical utility. HelixDock demonstrates outstanding capabilities on both cross-docking and structure-based virtual screening benchmarks. This study illuminates the strategic advantage of leveraging a vast and varied repository of generated data by physics-based tools to advance the frontiers of AI-driven drug discovery.
\end{abstract}

\keywords{Protein-ligand structure prediction \and Large-scale docking dataset \and Scaling laws \and Drug discovery}

\section{Introduction}
Protein-ligand structure prediction, a computational technique central to drug discovery, predicts the binding poses between small molecules (ligands) and target proteins (receptors). Recognized for its pivotal role in helping scientists identify potential drug candidates efficiently, the demand for improved accuracy in complex structure prediction has attracted substantial research attention in recent years.

The protein-ligand structure prediction technique has proceeded along two principal paths that focus on either the physics-based methods or the deep learning-based model.  The conventionally used physics-based docking tools, such as LeDock \cite{zhao2013ledock}, AutoDock \cite{trott2010autodock}, AutoDock Vina \cite{doi:10.1021/acs.jcim.1c00203}, Smina \cite{koes2013smina}, and Glide \cite{friesner2004glide,halgren2004glide}, are based on physics-based force fields and take into consideration various factors, such as shape complementary, electrostatics, hydrogen bonding, and van der Waals forces, to produce the candidate binding poses of the given receptor and ligand. These tools apply various sampling techniques to explore the conformational space and evaluate the sampled poses by the scoring functions. Despite the sound theoretical basis, this approach still suffers high challenges due to the limited conformational sampling and imprecise scoring. Recently, deep learning-based methods like EquiBind \cite{stark2022equibind}, TankBind \cite{lu2022tankbind}, DiffDock \cite{corso2022diffdock}, and Uni-Mol \cite{zhou2023uni} has emerged as alternatives, leveraging known complex structures to train models and potentially surpass traditional physics-based tools in prediction accuracy. However,  due to the high cost of experimentally determining receptor-ligand complex structures, there is very limited data available for training. As mentioned in the work \cite{crossdocked}, the performance of these models is susceptible to being over-optimistic, leading to concerns about their generalization capabilities, particularly when faced with novel complexes (i.e. dissimilar to those in the training set).

Here we reasoned that an improved protein-ligand structure prediction model could be developed by pre-training the deep learning-based structure prediction model on large-scale data generated by the physics-based docking tool, which hopefully can combine the complementary advantages of these two approaches.  Pre-training a model on a large and diverse dataset has demonstrated efficacy in various applications, particularly natural language processing and computer vision, to improve the accuracy of the models \cite{erhan2010does,he2019rethinking,dong2019unified}.  Recent endeavors in life sciences have sought to understand small molecules and proteins by exploiting large-scale data. Specifically, some studies \cite{DBLP:conf/iclr/HuLGZLPL20,DBLP:conf/nips/RongBXX0HH20,fang2022geometry,zhou2023uni,chen2023structure,li2022geomgcl,zhang2022helixadmet} learn the molecular representations from large-scale molecular databases that can be used for a variety of property prediction tasks, such as toxicity and binding affinity. There are also some studies \cite{rao2019evaluating,rives2019biological,rao2021msa,lin2023evolutionary,weissenow2022protein,fang2023method} to learn protein representation from a huge amount of protein sequences, which are then utilized for attribute annotation and structure prediction. In particular, Uni-Mol \cite{zhou2023uni} trained a molecular model and a protein pocket model, independently, for protein-ligand structure prediction. However, the interaction dynamics between protein receptors and molecule ligands are often overlooked due to the limited number of known receptor-ligand binding conformations. The PDBbind database, despite being a widely acknowledged comprehensive receptor-ligand complex repository, contains only around 20,000 experimentally validated structures, highlighting the data scarcity challenge. Though several molecular docking datasets with a great number of molecules, e.g., Cross-docking benchmark \cite{wierbowski2020cross} and a SARS-CoV2 dataset \cite{rogers2023sars}, can supplement the data, these datasets predominantly focus on a very limited number of frequently studied protein targets, leading to potential biases in model training. In contrast, pre-training on an extensive and diverse dataset generated by physics-based docking tools could endow a deep learning model with the requisite physical knowledge that is already considered by physics-based docking tools, thereby enhancing both precision and generalizability.

\begin{figure*}[h]
\centering
  \centering
  \includegraphics[width=1.0\linewidth]{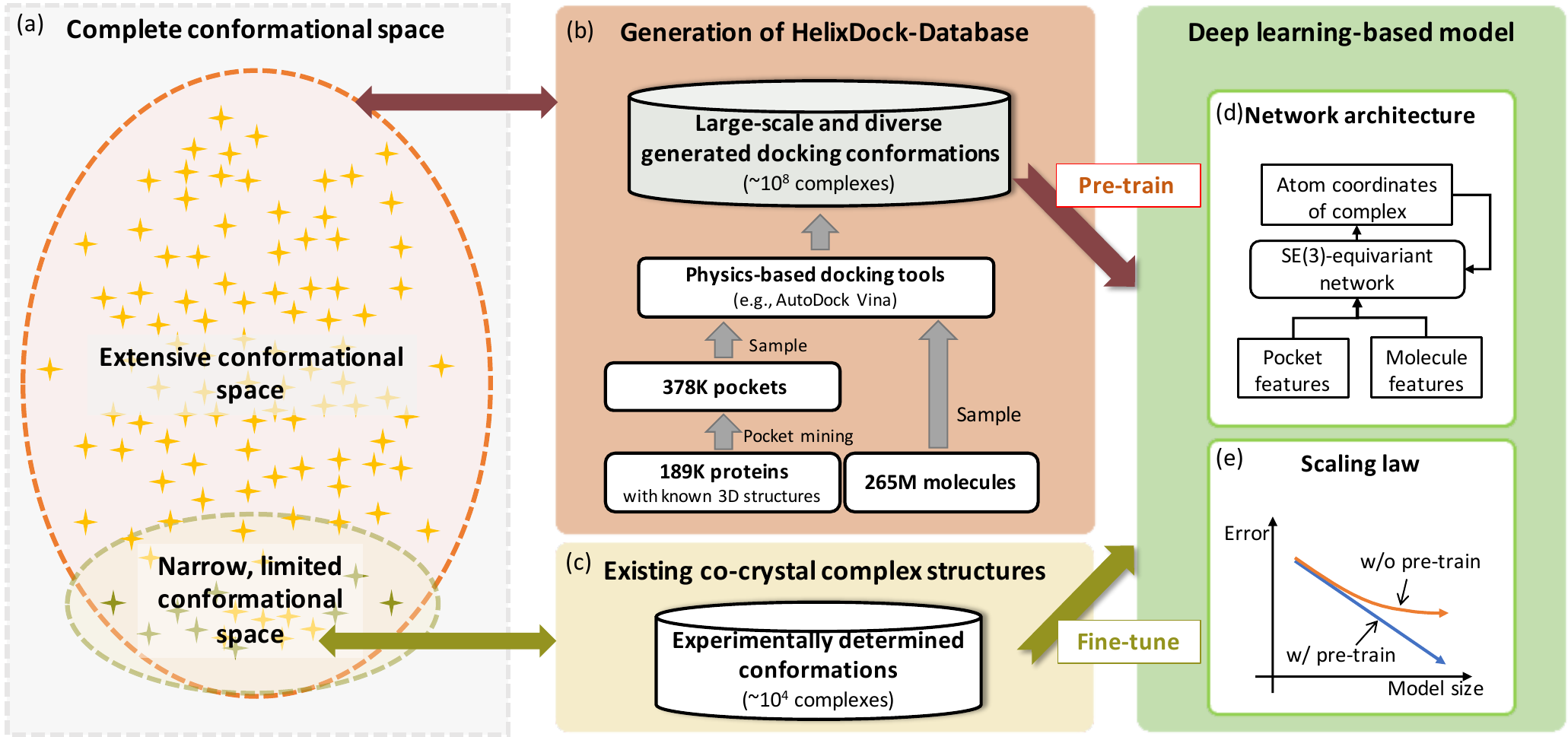}
  \caption{Overall framework of HelixDock: deep learning-based protein-ligand structure prediction model enhanced by massive and diverse binding poses.}
  \label{fig:helixdock_overall_framework}
\end{figure*}

Our proposed solution, HelixDock pre-trained a SE(3)-Equivariant network on an extensive collection of docking conformations generated by conventional physics-based tools, and fine-tuned with experimentally verified receptor-ligand complexes. HelixDock focuses on resolving the site-specific protein-ligand structure prediction in this study, and small molecules are docked into pre-identified binding pockets, usually identified through experimental techniques or expertise about the protein target. We generated 100 million binding poses using traditional physics-based molecular docking tools which is a task consuming roughly 1 million CPU core days.  The generated dataset contains hundreds of thousands of protein targets with known structures and millions of drug-like small molecules (as shown in Figure \ref{fig:helixdock_overall_framework}b). The generated poses are distributed in the extensive conformational space, as illustrated in Figure \ref{fig:helixdock_overall_framework}a.  These binding poses were then utilized for pre-training our designed SE(3)-Equivariant network model (Figure \ref{fig:helixdock_overall_framework}d), which serves as the foundation for learning general knowledge for complex structure prediction. Then, a small number of precise receptor-ligand complex structures, which are distributed in the narrow, limited conformational space (as illustrated in Figure \ref{fig:helixdock_overall_framework}a), were leveraged to fine-tune the model (as illustrated in Figure \ref{fig:helixdock_overall_framework}c). While the quality of the binding poses generated by the traditional physics-based docking tools may not match that of experimentally determined complex structures, we believe that our model can still distill the physical interaction knowledge between receptors and ligands from a large-scale dataset of rough binding poses.

Our comparative analyses have benchmarked HelixDock against various strong baseline methods, including the physics-based docking tools and the deep learning-based models, showing outstanding performance. HelixDock exhibits a comprehensive performance advantage over the baseline methods on multiple test sets. \fang{To be specific, HelixDock gives highly accurate predictions in two popular benchmarks, i.e., the PDBbind core set and PoseBusters Benchmark. Furthermore, HelixDock's superior performance on protein targets with low sequence identity to the training set underscores its robust transferability in protein-ligand structure prediction.}

Notably, we conduct training on multiple versions of structure prediction models, each with varying numbers of parameters and diverse pre-training data quantities, aiming to uncover the scaling laws governing the performance of deep learning-based structure prediction models. The investigation into the scaling laws \cite{hestness2017deep,kaplan2020scaling,alabdulmohsin2022revisiting} of the size of the training dataset and the number of model parameters can provide a rough estimation of the capabilities of the deep protein-ligand structure prediction models.   We experimentally show that pre-trained models consistently demonstrate performance improvement with an increase in model parameters, while non-pre-trained models struggle to achieve significant breakthroughs in performance with parameter increments (as shown in Figure \ref{fig:helixdock_overall_framework}e). 

\fang{
We sought to validate the practical applicability of HelixDock by applying it to two key tasks in drug discovery: cross-docking and structure-based virtual screening. In cross-docking, HelixDock maintains a high success rate across two datasets, outperforming other baselines. Moreover, in structure-based virtual screening, HelixDock demonstrated a remarkably high Enrichment Factor (EF) on the benchmark DUD-E dataset \cite{mysinger2012dude}, which encompasses 102 targets. } Our study showcases the potential of using large-scale and diverse receptor-ligand docking poses for pre-training deep learning-based docking models, which can shed light on future investigation with vast and varied generated data by physics-based tools in advancing AI-driven drug discovery.

\section{Methodology}

\subsection{Generation of Large-scale Docking Complexes}
\label{sec:helixdock_database}

\begin{table*}[ht]
    \centering
    \begin{tabular}{c|ccccccc}
    \toprule
    Dataset & \#Protein & \#Protein Family & \#Complex & Structure Source \\
    \midrule
    PDBbind version 2020 & 19k & 661 & 19k & Experimentally determined \\
    HelixDock-Database & 189k & 2,589 & 100m & Generated by AutoDock Vina \\
    \bottomrule
    \end{tabular}
    \caption{Statistics of HelixDock-Database and PDBbind.}
    \label{tab:dataset_summary}
\end{table*}

\begin{figure*}[!htbp]

  \centering
  \includegraphics[width=1.0\linewidth]{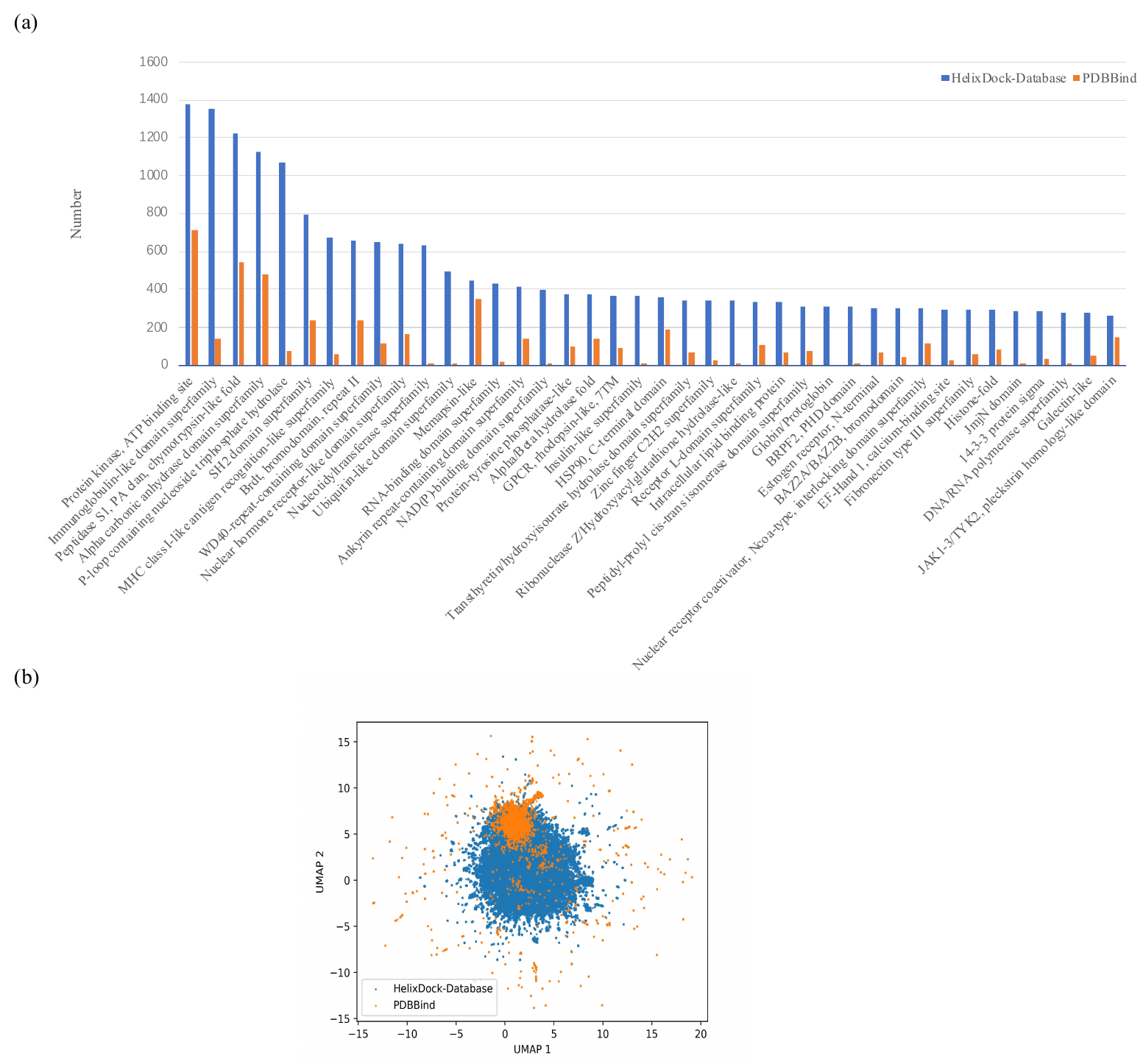}
\caption{Comparison of HelixDock-Database and PDBbind.(a) Protein family comparison of HelixDock-Database and PDBbind. (b) UMAP of Morgan fingerprints of ligands from HelixDock-Database and PDBbind.}
\label{fig:dataset_compare}
\end{figure*}

HelixDock utilizes a vast and varied array of docking complexes generated through physics-based docking tools, AutoDock Vina \cite{trott2010autodock}, to enhance the accuracy of protein-ligand structure prediction. As shown in Figure \ref{fig:helixdock_overall_framework}, our methodology first involves the generation of poses of approximately 100 million receptor-ligand pairs utilizing a supercomputing platform.
These systematically generated docking complexes function as the foundational pre-training dataset for HelixDock, allowing for an intricate understanding and learning of the underlying physical principles involved in molecular interactions. This dataset is herein referred to as the HelixDock-Database. The construction of the HelixDock-Database consists of three phases.

In the first phase, we establish the protein target set by consolidating all proteins with known 3D structures from the Protein Data Bank (PDB) \cite{berman2000proteindatabank} available until September 31, 2022, encompassing approximately 189 thousand proteins. These proteins manifest a diverse array of protein families, encapsulating a total of 2,589 distinct families. In comparison, the widely acknowledged receptor-ligand complex dataset, PDBbind \cite{wang2004pdbbind}, incorporates only 19 thousand proteins and encompasses only 661 protein families, as shown in Table \ref{tab:dataset_summary}. The numeric distributions among the top-40 protein families within the HelixDock-Database are illustrated in Figure \ref{fig:dataset_compare}a, and are concurrently contrasted with the distributions of protein families within PDBbind. Subsequently, fpocket \cite{le2009fpocket}, an advanced pocket mining tool, is employed to determine potential binding pockets within each protein to streamline the subsequent docking procedures. The two most probable pockets from each protein were selected,  yielding approximately 378 thousand candidate target pockets.

We curate the ligand set in the second phase, acquiring 265 million drug-like small molecules from Enamine (enamine.net), a compound library renowned for its prevalent application in virtual screening endeavors. The selection of these ligands is strategically conducted to ensure they encompass extensive coverage in the drug-like hit space. 
Figure \ref{fig:dataset_compare}b contrasts the chemical space represented by 1 million ligands randomly selected from the HelixDock-Database, against the chemical space represented by the entire set of 19,000 ligands from PDBbind. This comparative visualization employs the UMAP algorithm \cite{mcinnes2018umap} to project the Morgan fingerprints of these ligands into a two-dimensional representation. As shown in Figure \ref{fig:dataset_compare}b, the ligands within the HelixDock database have more expansive coverage of the chemical space compared to those within PDBbind.

In the final phase, we execute a comprehensive large-scale docking initiative, pairing targets from the pre-established pocket target set with ligands from the curated ligand set through a random selection process, as illustrated in Figure \ref{fig:helixdock_overall_framework}b. To clarify, one target pocket and one compound are randomly sampled to constitute a target-ligand pair, leading to the generation of approximately 100 million such pairs. AutoDock Vina \cite{trott2010autodock}, a widely recognized docking software, is employed for generating docking poses. 
To maintain a balanced representation of protein targets, we perform clustering of proteins based on sequence similarity with the MMSeqs2 tool \cite{steinegger2017mmseqs2}. The likelihood of each protein being sampled is calculated to be inversely proportional to the size of its corresponding protein cluster, ensuring the representation of a diverse range of protein targets. This methodology affirms the diversity within our protein targets. The docking of all these 100 million pairs takes approximately 1 million CPU core days to complete.

The coarse-grained conformations derived through this methodology serve as foundational elements for the pre-training of the deep learning-based model. The precision of these conformations, as crafted by docking tools, does not align with the accuracy found in experimentally derived receptor-ligand complex conformations. Nonetheless, we postulate that a molecular docking tool holds the potential to elucidate the fundamental physical principles that govern docking interactions. Typically, physics-based docking tools usually traverse conformational space and scrutinize conformations leveraging sophisticated physics-based scoring functions.

\subsection{Training Paradism and Network Architecture of HelixDock}
\label{sec:helixdock_network}

HelixDock's training process consists of two main phases: a pre-training phase and a fine-tuning phase. Initially, in the pre-training phase, the model is trained using the HelixDock-Database. This database provides large-scale data that helps the model learn the general features and patterns of protein-ligand structures. Leveraging this extensive dataset enables the model to gain a broad understanding and robust foundation for structure prediction. In the fine-tuning phase, the model is further refined using the general set from PDBbind version 2020 \cite{wang2004pdbbind}. This set includes experimentally derived co-crystal conformations, which provide high-quality and specific structural information. To ensure unbiased fine-tuning, the core set is excluded from this phase. This two-step training process allows HelixDock to combine the strengths of large-scale pre-training with targeted fine-tuning, resulting in a highly accurate and reliable protein-ligand structure prediction model.

\lihanga{
HelixDock improves protein-ligand conformation prediction accuracy by leveraging geometry-aware neural network architectures and adopting a molecular diffusion process.
Following the recent success of using the diffusion model to learn continuous distributions \cite{ho2020ddpm}, we generate train inputs by adding Gaussian noise into the 3D coordinates of all heavy atoms in the ligand, then HelixDock is trained to reverse the noising process by predicting the original docking/crystal conformations, instead of predicting the injected noise in the original paper. More concretely, for each time step of the diffusion process,
}
HelixDock directly predicts the 3D coordinates of all heavy atoms in the ligand based on the information of the protein pockets (with known structures) and the ligands (with \lihanga{noised} structures). The architecture of HelixDock consists of two main modules: InteractionLearner which utilizes a hybrid of graph neural networks (GNNs) and the transformer-style $\text{GeoFormer}$ network, focusing on modeling interactions between protein pockets and ligands; and StructurePredictor which adopts an end-to-end SE(3)-Equivariant network to predict and iteratively refine the 3D coordinates of the atoms in the ligands. 

\fang{When utilizing HelixDock to predict protein-ligand conformations, we randomly sample multiple noise inputs to generate a diverse array of conformations. This broad exploration ensures various potential binding poses. Similar to traditional docking tools and the DiffDock method, we use a scoring function to evaluate and rank these conformations. The scoring function assesses predicted binding affinity and physical plausibility, allowing us to identify the most accurate and biologically relevant conformations.}

\section{Pre-training on Large-scale Docking Complexes Enhances Complex Structure Prediction}
\label{section:exp_pose}

\begin{figure*}[!htbp]

  \centering
  \includegraphics[width=0.9\linewidth]{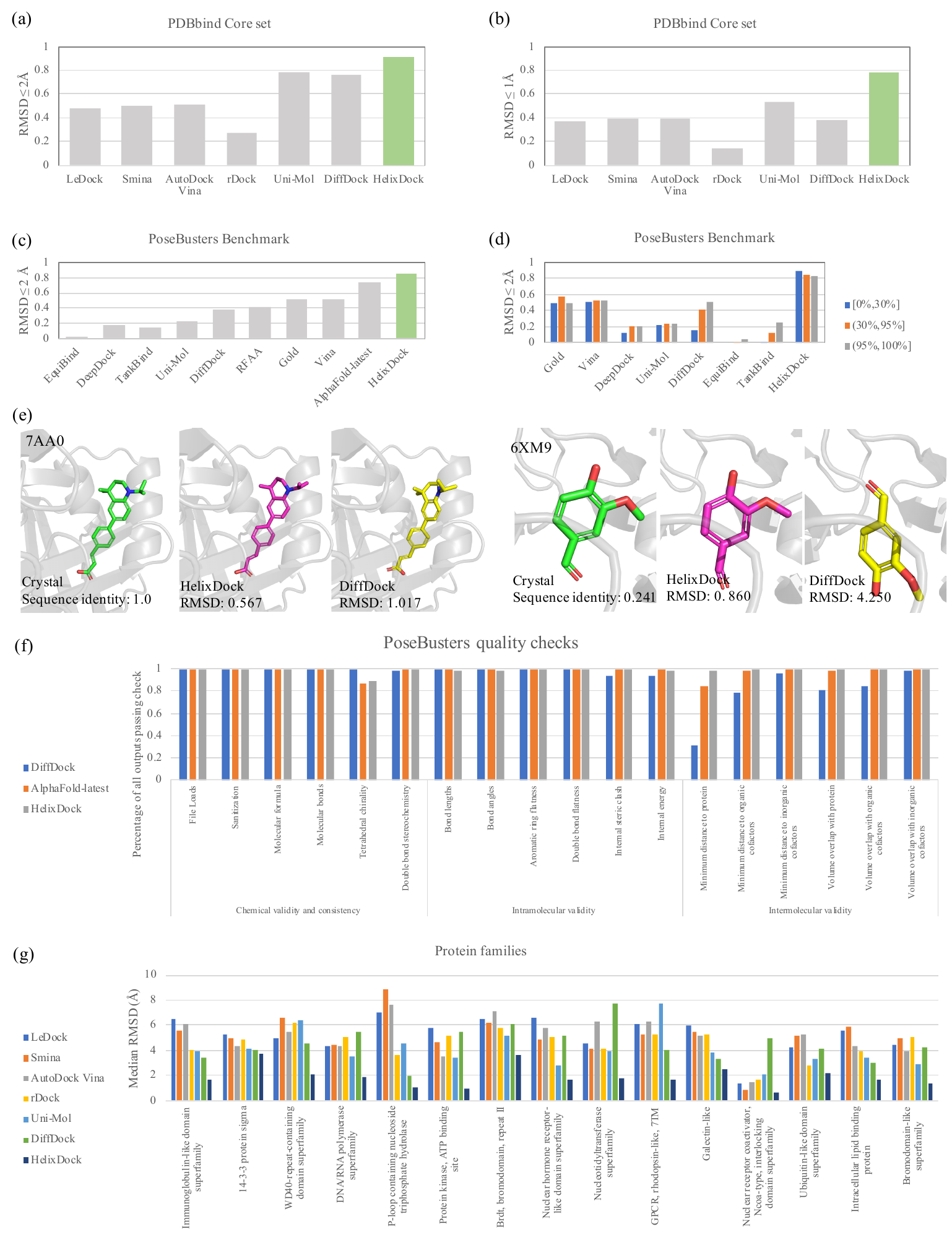}
\caption{\lihanga{Overall evaluation of HelixDock and the baseline methods for complex structure prediction on complex prediction datasets. (a) Percentage of RMSD $\leq 2$\AA~(success rate) on PDBbind Core set. (b) Percentage of RMSD $\leq 1$\AA~on PDBbind Core set. (c) Percentage of RMSD $\leq 2$\AA~on PoseBusters benchmark. (d) Success rate in PoseBusters benchmark with cases stratified by sequence identity to the PDBbind 2020 General set. (e) Two examples from PoseBusters, PDB\_id:7AA0 with a high sequence identity of 1.0 and PDB\_id:6XM9 with a low sequence identity of 0.24. (f) PoseBuesters quality checks on predictions of DiffDock, AlphaFold-latest, and HelixDock. (g) Performance comparison across various protein families.}}
\label{fig:overall_exp}
\end{figure*}

\lihanga{
\subsection{Superior Performance of HelixDock Compared to Baseline Methods}
We began our evaluation by examining the overall performance of HelixDock. The overall performance of HelixDock in predicting complex structures is evaluated on PDBbind core set \cite{su2018comparative} and PoseBusters benchmark \cite{buttenschoen2024posebusters}.
We compared HelixDock against two categories of protein-ligand structure prediction methods: the physics-based docking tools including LeDock \cite{zhang2016enriching}, Smina \cite{koes2013smina}, AutoDock Vina \cite{trott2010autodock}, and rDock \cite{ruiz2014rdock}, as well as deep learning-based protein-ligand structure prediction models, including Uni-Mol \cite{zhou2023unimol} and DiffDock \cite{corso2022diffdock}. In accordance with prior research \cite{wang2016comprehensive,stark2022equibind},  we employ the Root Mean Square Deviation (RMSD) as our evaluation metric, comparing the predicted ligand poses with their respective native ligand binding poses, where only heavy atoms are taken into consideration.

Figure \ref{fig:overall_exp}a\&b present the performance of HelixDock and other baseline methods on the PDBbind core set. HelixDock achieves a high success rate (RMSD $\leq 2$\AA) of 90.1\%. This marks a significant improvement over the second-best method under the same metric. HelixDock's performance is particularly distinguished at the percentage of RMSD $\leq 1$\AA, surpassing the top baseline method by over 20\%. This significant improvement highlights HelixDock's exceptional capability in predicting conformational details with high precision.
}

\lihanga{
\subsection{HelixDock demonstrates Its Transferability on Novel Complexes}
Even though the deep-learning-based baselines exhibit an obvious advantage over physics-based methods on the PDBbind core set, previous research \cite{crossdocked} has indicated that deep-learning models trained on the general set are susceptible to overestimating their performance on the core set given the observed similarities between the samples in the PDBbind core set and PDBbind general set. The transferability of deep learning methods to novel complexes, particularly novel protein targets, is a key concern.

To this end, we further evaluated our model using the recently developed PoseBusters Benchmark set, a high-quality collection of protein-ligand complexes \cite{buttenschoen2024posebusters}. This set comprises 428 complexes introduced since 2021, with most exhibiting low similarity to samples in our training dataset. HelixDock achieves a high success rate of 85.6\% in structural predictions, as depicted in Figure \ref{fig:overall_exp}c. This performance surpasses that of the baseline methods, which are detailed in the records from PoseBusters and AlphaFold-latest \cite{af-latest}. Notably, while other deep-learning-based methods generally fall short compared to physics-based approaches, HelixDock and AlphaFold-latest stand out as exceptions, consistently delivering superior results.

To further assess the transferability of HelixDock to novel protein targets, we stratified the PoseBusters Benchmark set according to the target protein receptor’s maximum sequence identity with proteins in the PDBbind 2020 General Set \cite{wang2004pdbbind}, following the methodology described in \cite{buttenschoen2024posebusters}. The test cases are categorized into three groups based on maximum percentage sequence identity: low [0\%, 30\%], medium (30\%, 90\%], and high (90\%, 100\%]. As shown in Figure \ref{fig:overall_exp}d, physics-based methods exhibit consistent performance across all three groups, while previous deep-learning-based methods show a decline in performance with proteins of lower sequence identity. For instance, DiffDock's success rate dramatically drops from 51\% in the high sequence identity category to just 16\% in the low sequence identity category. In contrast, HelixDock maintains consistent performance across these groups, exemplifying its robust generalization capability to new protein targets. 
As depicted in Figure \ref{fig:overall_exp}e, 
HelixDock exhibits robust docking performance across complexes from the PoseBusters dataset, achieving high accuracy irrespective of the sequence similarity with PDBbind 2020 general set, whether at 1.0 or 0.24.

\subsection{HelixDock Produces Physically Plausible Molecular Structures}
Unlike physics-based methods, deep learning-based conformation prediction methods are not guaranteed to produce physically plausible molecular structures, which is one of the main reasons these methods are frequently criticized. To evaluate the chemical and geometric consistency of the predicted ligand conformations, the PoseBusters test suite \cite{buttenschoen2024posebusters} has implemented a series of standard quality checks. These checks assess the stereochemistry of the ligands as well as the physical plausibility of intra- and intermolecular measurements. Ligand conformations that successfully pass all these tests are deemed ‘PB-valid’.

The comparison of HelixDock with DiffDock and AlphaFold-latest on these quality checks is shown in Figure \ref{fig:overall_exp}f. Overall, HelixDock achieves a PB-valid of 85.2\%. It surpasses or matches DiffDock in 15 out of 18 quality checks and outperforms or equals AlphaFold-latest in 12 of these checks. While HelixDock does not guarantee that all generated conformations are physically plausible, its overall performance in producing valid structures is commendable.

\subsection{HelixDock Maintains High Accurate Predictions Across Various Protein Families}
We extracted complexes released between January 1, 2020, and December 31, 2022, from the RCSB PDB \footnote{https://rcsb.org} and categorized the protein targets according to their protein families. Figure \ref{fig:overall_exp}g displays the results for the top 15 protein families.
Notably, HelixDock achieves the best performance among all considered protein families, recording the lowest RMSD values. It further distinguishes itself by achieving a median RMSD value below 2\AA~ across 10 protein families.
These outcomes validate the robustness of HelixDock across a wide spectrum of protein families.
}
This evaluation reaffirms HelixDock's promise as a reliable tool for protein-ligand structure prediction studies, exhibiting excellent prediction with lower RMSD values and adaptability to diverse protein families.

\section{Scaling Law for Protein-Ligand Structure Prediction}
\label{sec:scaling_law}

\begin{figure*}
  \centering
  \includegraphics[width=1.0\linewidth]{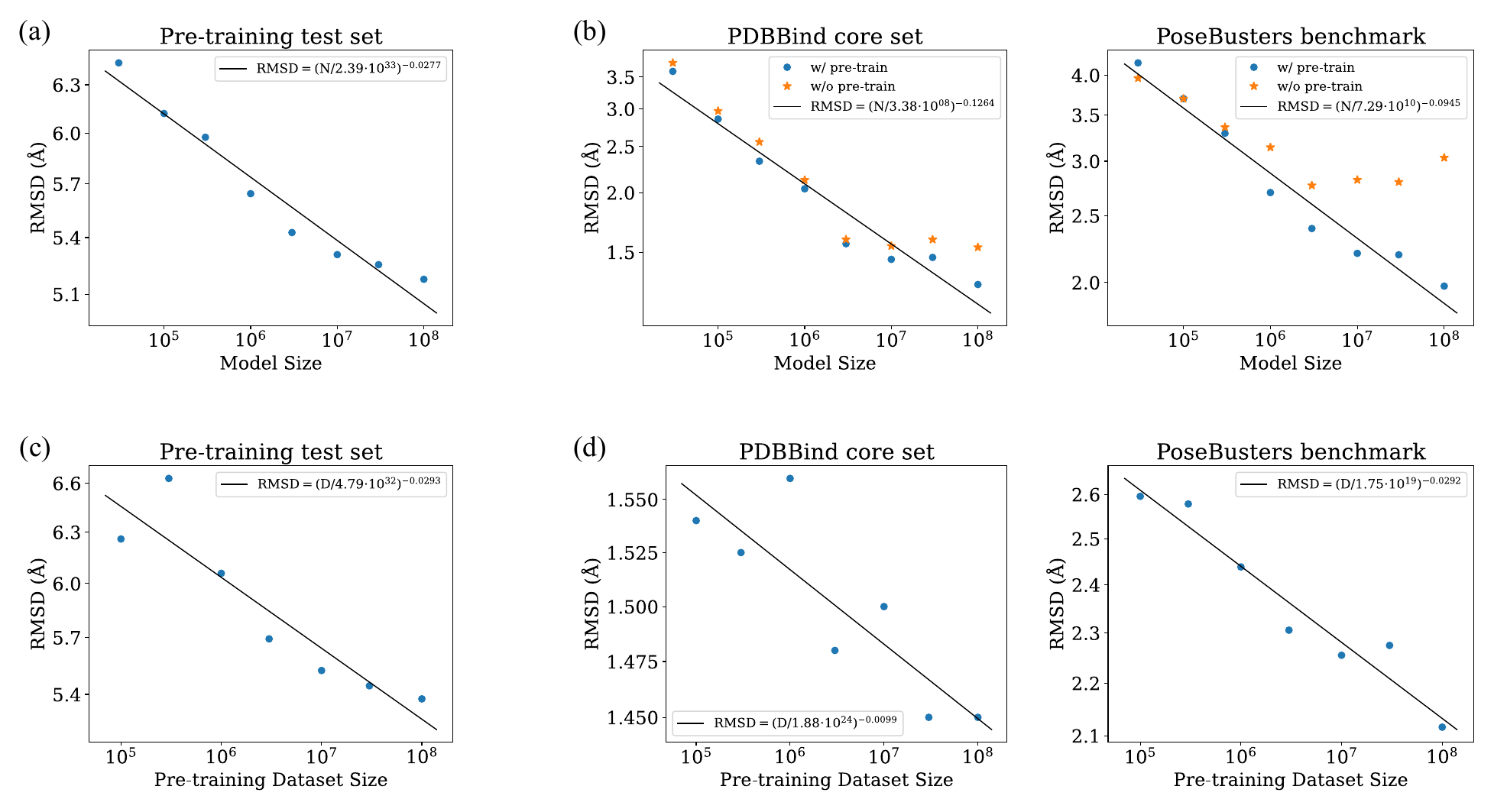}
\caption{Scaling laws for protein-ligand structure prediction. (a) Scaling laws of model sizes in the pre-training stage. (b) Scaling laws of model sizes in the fine-tuning stage with models evaluated on PDBbind core set and PoseBusters benchmark. (c) Scaling laws of pre-training data sizes in the pre-training stage. (d) Scaling laws of pre-training data in the fine-tuning stage with models evaluated on PDBbind core set and PoseBusters benchmark.}
\label{fig:scaling_law}
\end{figure*}

Recent advances in Natural Language Processing and Computer Vision have highlighted the phenomenon of empirical scaling laws, where enlarging both the data size and the model size can substantially boost model performance across a variety of applications \cite{hestness2017deep,kaplan2020scaling,alabdulmohsin2022revisiting,ghorbani2021scaling,zhai2022scaling}.  In this section, we seek to explore the effects of model size $N$, characterized by the number of model parameters, and pre-training dataset size $D$ in the domain of protein-ligand structure prediction. Our objective is to experimentally investigate the empirical scaling laws in this specific application. We hope this analysis can offer valuable insights for our future work and that of the broader research community.

In our evaluation, we investigate the empirical scaling laws at both the pre-training and fine-tuning phases of our model. In the pre-training phase, we present the average RMSD value for pre-trained models. In order to evaluate the pre-trained models, we constructed a held-out pre-training test set, which is a randomly chosen subset from the HelixDock-Database, comprising 36,000 samples with 180 distinct proteins. In the fine-tuning phase, we report the average RMSD values for models refined using PDBbind, evaluated on the PDBbind core set \lihanga{and PoseBusters benchmark}. For all model variations subjected to pre-training, we only conduct the pre-training for a fixed duration of 120,000 steps due to computational resource constraints. In the fine-tuning phase, we perform three independent runs for each model to reduce variance, presenting the average results.

From our experimental results, it is evident that both the model size and the pre-training dataset size play a crucial role in enhancing the predictive accuracy of the complex structure prediction task. The scaling laws previously validated in other fields remain effective in protein-ligand structure prediction.

\subsection{Relations between the Performance and the Model Sizes}

We begin by studying the relation between the model size (ranging from $3\times 10^4$ to $10^8$) and the RMSD values of these test sets during the pre-training and the fine-tuning stage. Notably, when increasing the model size $N$, we keep the pre-training size $D=10^{8}$ fixed for all the variants. We also assess the models without pre-training to analyze the impact of pre-training for complex structure prediction. In line with prior research\cite{ghorbani2021scaling}, we express RMSD as a function of model size using the power-law relationship described in Equation \ref{eq:scaling_model}:
\begin{equation}
    \text{RMSD}(N) \approx (N / N_c)^{\alpha_N}. \label{eq:scaling_model}
\end{equation}
Here, $N_c$ and $\alpha_N$ are constant parameters of the power-law that we fitted to the data points. $\alpha_N$ provides insights into the degree of performance improvement to some extent.

The results of the pre-training stage and the fine-tuning stage are depicted in Figure \ref{fig:scaling_law}a and Figure \ref{fig:scaling_law}b, respectively. Our results strongly suggest a robust correlation between the accuracy of structure prediction models and their respective model sizes during both the pre-training and fine-tuning stages. An interesting observation is that when the pre-training technique is not involved (the w/o pre-train variants i

n Figure \ref{fig:scaling_law}b), it becomes increasingly challenging to achieve additional gains by scaling up the model on test sets of both the PDBbind core set (after $N=3\times 10^{6}$) and \lihanga{and PoseBusters (after $N=3\times 10^{6}$)}. On the other hand, when pre-training (the w/ pre-train variants in Figure \ref{fig:scaling_law}b) is employed, scaling up the model consistently leads to continued improvements in performance. These relationships between $N$ and RMSD in both stages can be closely approximated by a power law as shown in Equation \ref{eq:scaling_model}.
Moreover, as the model size grows, we observe a more substantial performance gap between pre-trained and non-pre-trained variants in the PoseBusters benchmark compared to the PDBbind core set. This phenomenon might be attributed to the greater complexity of cases in the \lihanga{PoseBusters benchmark}.
This finding underscores the pivotal role of pre-training in augmenting model scalability for protein-ligand structure prediction.

\subsection{Relations between the Performance and the Pre-training Data Sizes}

In this section, we validate one of our primary contributions through empirical scaling law experiments: the generation of an extensive collection of docking complexes and the utilization of these docking complexes for pre-training purposes. Specifically, we vary the pre-training data size from $10^5$ to $10^8$ and evaluate the performance of the respective pre-trained models. The model size is kept fixed at $N=10^{7}$. Similar to the analysis of model size, we use a power-law function to fit the relationship between RMSD values and the pre-training dataset size:

\begin{equation}
    \text{RMSD}(D) \approx (D/D_c)^{\alpha_D}.
\end{equation}

The results are illustrated in Figure \ref{fig:scaling_law}c and Figure \ref{fig:scaling_law}d. There is a discernible correlation between the pre-training dataset size ($D$) and the performance of the model (measured by RMSD). These findings suggest a potential for further enhancement in the structure prediction model's performance with increasing pre-training dataset sizes, demonstrating the substantial benefits of the large-scale pre-training docking complexes.

\lihanga{
\section{Practicality in Drug Discovery}
To validate the practical utility of HelixDock in drug development, we first apply it to cross-docking (i.e., unbound docking) and structure-based virtual screening, evaluating its performance in these critical tasks.

\subsection{HelixDock Generalizes Well to Cross-Docking}

\begin{figure*}[!htbp]
  \centering
  \includegraphics[width=1.0\linewidth]{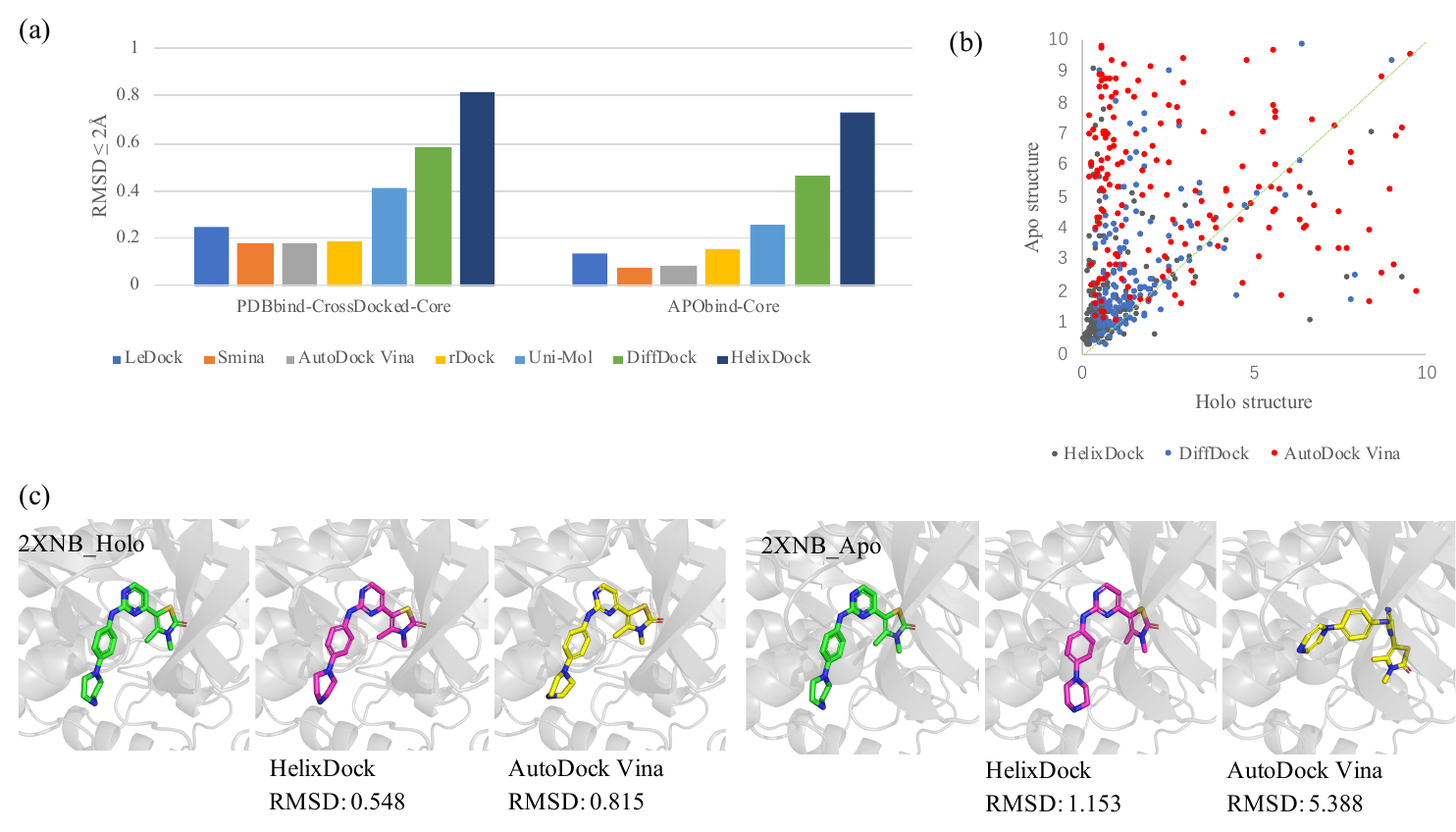}
\caption{\lihanga{Cross-docking. (a) Performance of HelixDock and other baselines in two cross-docking datasets, i.e., PDBbind-CrossDocked-Core and APObind-Core. (b) RMSD comparison of HelixDock on holo structures and apo structures. (c) Prediction of HelixDock on sample PDB ID:2XNB from PDBbind Core set when the protein is holo or apo.}}
\label{fig:cross-docking}
\end{figure*}

The preceding experiments assumed the receptor structure to be in its optimal conformation, but this does not align with real-world scenarios in protein-ligand structure prediction, where receptor structures exhibit flexibility. Cross-docking involves extracting a ligand from a co-crystal complex and docking it to a different conformation (apo structure) of the same protein, rather than to the ligand's original holo structure. We evaluate HelixDock on two cross-docking datasets, the PDBbind-CrossDocked-Core \cite{shen2021crossdock} including 1,058 cross-docked complexes and APObind-Core \cite{aggarwal2021apobind} including 229 cross-docked complexes. 

As illustrated in Figure \ref{fig:cross-docking}a, HelixDock continues to achieve a high success rate in both the PDBbind-CrossDocked-Core and APObind-Core datasets, reaching 81.7\% and 72.9\%, respectively. In contrast, the performance of baseline methods on these cross-docking datasets falls significantly short. This drop in performance is especially pronounced when compared to their effectiveness in the re-docking dataset, where receptor and ligand structures are more compatible. This decline underlines the difficulty of predicting complex structures when introduced to alternative protein conformations, a challenge that HelixDock seems particularly well-equipped to overcome. 

In Figure \ref{fig:cross-docking}b, we conducted a comprehensive analysis of the RMSDs associated with each protein-ligand pair predicted by HelixDock, DiffDock, and AutoDock Vina, considering both holo and apo structures. Across all methodologies, a notable trend emerged wherein numerous samples displayed decreased efficacy when the target protein was in the apo state as opposed to the holo state. Notably, HelixDock demonstrated a comparatively lesser degree of influence. The majority of these predictions maintained a commendable level of accuracy despite the structural variations encountered.
For example, as illustrated in Figure \ref{fig:cross-docking}c, both HelixDock and AutoDock Vina demonstrate strong performance on the 2XNB complex in its holo structure. However, in the apo structure, HelixDock maintains a low RMSD of 1.153 \AA~, whereas the RMSD of AutoDock Vina significantly rises to 5.388\AA~.

}

\subsection{HelixDock Demonstrates Application Potential in Structure-Based Virtual Screening}

\begin{figure*}[!htbp]
  \centering
  \includegraphics[width=0.85\linewidth]{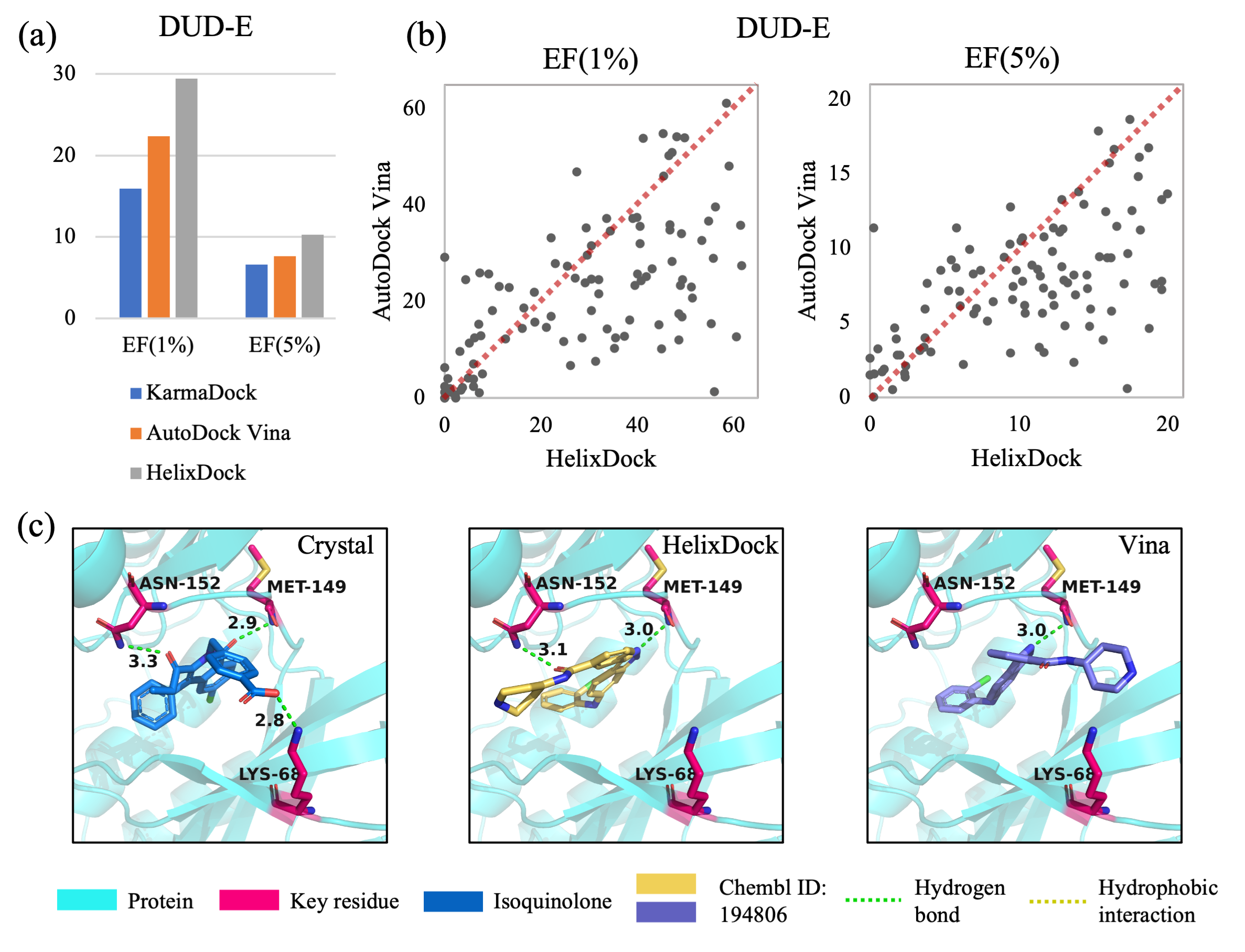}
\caption{Performance of HelixDock on structure-based virtual screening tasks and experimental validation. (a) Overall comparison of HelixDock and baseline methods on a virtual screening benchmark DUD-E. (b) Detailed comparison of HelixDock and AutoDock Vina on all the 102 targets in DUD-E. (c) The co-crystal pose of JNK3 binding with Isoquinolone (left, PDB code 2ZDT \cite{asanoDiscoverySynthesisBiological2008}), and binding poses with \href{https://www.ebi.ac.uk/chembl/compound_report_card/CHEMBL194806/}{ChEMBL194806} predicted by HelixDock (middle) and Vina (right), respectively.}
\label{fig:virtual-screening}
\end{figure*}

Virtual screening is a pivotal technique in drug discovery, employed to identify promising drug candidates from large-scale compound libraries. Typical structure-based virtual screening approaches involve generating the ligand conformation bound to the protein target and subsequently applying a scoring function to assess the binding strength. The compounds with the highest scores are selected for further validation. The accuracy of these binding poses significantly influences the final outcomes \cite{liontaStructureBasedVirtualScreening2014}.

To assess the efficacy of HelixDock, we compared its virtual screening capabilities with AutoDock Vina, a widely-used docking tool, and KarmaDock \cite{zhang2023karmadock}, an advanced AI-based virtual screening method. We evaluated these methods using the enhancement factor (EF) \cite{li2018assessing,su2018comparative} on the DUD-E dataset \cite{mysinger2012dude}, a widely adopted virtual screening benchmark comprising actives and decoys across 102 protein targets. As illustrated in Figure \ref{fig:virtual-screening}a, HelixDock surpasses both KarmaDock and AutoDock Vina in enrichment factors (EF) at 1\% and 5\% thresholds, underscoring its strong potential in real virtual screening applications. Specifically, HelixDock achieves an EF(1\%) of 29.4, which represents a significant 31.7\% improvement over AutoDock Vina's 22.3. Further analysis is presented in Figure \ref{fig:virtual-screening}b, where HelixDock and AutoDock Vina are compared across all protein targets in the DUD-E dataset. Here, HelixDock consistently outperforms AutoDock Vina in terms of EF across most targets, demonstrating its robustness and superior performance in virtual screening scenarios.

We also discovered that the binding poses generated by HelixDock demonstrate superior binding modes compared to those produced by other computational tools. As illustrated in Figure \ref{fig:virtual-screening}c, we visualized a protein target from the DUD-E database \cite{mysinger2012dude}, c-Jun N-terminal kinase 3 (JNK3), and compared the differences between the co-crystallized reference ligand's binding pose (ligand: isoquinolone, PDB code: 2ZDT) and the poses of another active compound (ChEMBL ID: 194806) generated by HelixDock and Vina. Previous research has highlighted a hydrogen bond between isoquinolone and the NH group of MET-149's main chain in the hinge region, as well as hydrogen bonds with ASN-152 and LYS-68, are crucial for the JNK3 inhibition \cite{asano2008discovery}. Our observations indicate that the binding pose predicted by HelixDock forms hydrogen bonds with both ASN-152 and MET-149, whereas the pose predicted by Vina establishes only one hydrogen bond with MET-149. The binding configuration predicted by HelixDock more accurately replicates the binding modes found in the co-crystal structure, which suggests that HelixDock is adept at capturing essential protein-ligand interactions, thereby providing a more reliable starting point for structure-based lead optimization.

\section{Conclusion and Future Work}
Protein-ligand structure prediction remains indispensable in drug discovery for its computational efficiency in predicting binding interactions. Recent strides in deep learning-based methods offer a more accurate alternative compared to physics-based docking, although they face inherent challenges in generalization due to limited available training data. HelixDock, however, capitalizes on a vast dataset of docking poses generated through physics-based tools, harnessing the potential of deep learning-based protein-ligand structure prediction.

Our findings demonstrate the superiority of HelixDock over baseline methods, particularly on challenging samples. Moreover, our study indicates that scaling up the pre-train data size can enhance performance. Besides, when applying pre-training, increasing the model size can lead to further performance improvement. In conclusion, our work not only highlights the substantial potential of large-scale data in life sciences but may also provide inspiration to fellow researchers in related fields.

Several promising directions for future research include:
\begin{itemize}
    \item Larger-scale data: We have generated hundreds of millions of molecular docking poses for pre-training and plan to expand this dataset, leveraging the advantages of scaling to enhance model precision.
    \item Higher-precision data: Recognizing the precision limitations of docking poses generated by physics-based methods, we explore more precise techniques like molecular dynamics simulations to optimize conformational data.
    \item Wider applications: Our ongoing efforts involve applying extensive conformational data to improve the performance of related tasks such as affinity prediction and molecular property estimation. Additionally, we consider the application of pre-training strategies to study large molecules, such as predicting protein complex structures.
\end{itemize}

\textbf{Code availability:} The source code, trained weights, and inference code of HelixDock will be freely available at GitHub (\url{https://github.com/PaddlePaddle/PaddleHelix/tree/dev/apps/molecular\_docking/helixdock}) to ensure the reproduction of our experimental results. A web service of HelixDock is also available at \url{https://paddlehelix.baidu.com/app/drug/helix-dock/forecast} to provide efficient structure predictions.

\clearpage

\bibliographystyle{unsrt}  
\bibliography{references}  

\end{document}